\documentclass[pdflatex,sn-basic]{sn-jnl}

\jyear{2023}%

\theoremstyle{thmstyleone}%
\theoremstyle{thmstyletwo}%

\theoremstyle{thmstylethree}%

\usepackage{times}
\usepackage{latexsym}
\usepackage{microtype}
\usepackage{graphicx}
\usepackage{subfigure}
\usepackage{amsmath}
\usepackage{amssymb}
\usepackage{threeparttable}
\usepackage{booktabs}
\usepackage{multirow}
\usepackage{pifont}
\usepackage{bm}
\usepackage{tabularx}

\usepackage{color}

\usepackage{lineno}

\begin{document}

\title[Accurate Use of Label Dependency in Multi-Label Text Classification]{Accurate Use of Label Dependency in Multi-Label Text Classification Through the Lens of Causality}


\author[1]{Caoyun Fan}\email{fcy3649@sjtu.edu.cn}
\equalcont{These authors contributed equally to this work.}
\author[2]{Wenqing Chen}\email{chenwq95@mail.sysu.edu.cn}
\equalcont{These authors contributed equally to this work.}
\author[1]{Jidong Tian}\email{frank92@sjtu.edu.cn}
\author[1]{Yitian Li}\email{yitian\_li@sjtu.edu.cn}
\author*[1]{Hao He}\email{hehao@sjtu.edu.cn}
\author[1]{Yaohui Jin}\email{jinyh@sjtu.edu.cn}

\affil*[1]{MoE Key Lab of Artificial Intelligence, AI Institute, Shanghai Jiao Tong University, Shanghai, China}
\affil[2]{School of Software Engineering, Sun Yat-sen University, Guangzhou, China}


\abstract{
Multi-Label Text Classification (MLTC) aims to assign the most relevant labels to each given text. Existing methods demonstrate that label dependency can help to improve the model's performance. However, the introduction of label dependency may cause the model to suffer from unwanted prediction bias. In this study, we attribute the bias to the model's misuse of label dependency, i.e., the model tends to utilize the correlation shortcut in label dependency rather than fusing text information and label dependency for prediction. Motivated by causal inference, we propose a CounterFactual Text Classifier (CFTC) to eliminate the correlation bias, and make causality-based predictions. Specifically, our CFTC first adopts the predict-then-modify backbone to extract precise label information embedded in label dependency, then blocks the correlation shortcut through the counterfactual de-bias technique with the help of the human causal graph. Experimental results on three datasets demonstrate that our CFTC significantly outperforms the baselines and effectively eliminates the correlation bias in datasets. }



\maketitle

\section{Introduction}
\label{s1}




Multi-Label Text Classification (MLTC) is a fundamental but challenging task \citep{DBLP:journals/apin/WangLLZZF20} in Natural Language Processing (NLP), and it has be applied in many real-world scenarios, such as text categorization \citep{DBLP:journals/tcss/WangCLG19}, sentiment analysis \citep{DBLP:journals/air/WankhadeRK22}, emotion recognition \citep{DBLP:journals/kais/AlswaidanM20} and so on. Therefore, it is necessary to design an accurate and efficient multi-label text classifier for practical applications. 

\begin{table}[h]
    \centering
    \caption{An example of MLTC. In this example, If some labels are known (listed in \emph{Given Labels}), we are more likely to predict some labels (listed in \emph{Positive Labels}) and exclude others (listed in \emph{Negative Labels}). }
    \fontsize{10}{10}\selectfont 
    \begin{threeparttable}
    \begin{tabular}{lc}
    \toprule
    \bf Text & Germany beat Argentina on Gotze's goal to win World Cup. \cr
    \midrule
        \bf Given Labels & Sport, World Cup \cr
        \bf Positive Labels & Football \cr
        \bf Negative Labels & Technology, Education, Economics \cr
    \bottomrule
    \end{tabular}
    \end{threeparttable}
    \vspace{-0.3cm}
    \label{tab1-1}
\end{table}

Initially, MLTC was decomposed into multiple independent binary classification tasks \citep{Boutell2004LearningMS}. Soon the researchers realized that Label Dependency (LD) \citep{DBLP:journals/apin/ChenR21} could be exploited to improve performance. Intuitively, knowing some labels makes it easier to predict other labels, because the labels tend to have dependency \citep{DBLP:conf/coling/YangSLMWW18}. Taking Table \ref{tab1-1} as an example, the given labels `Sports, World Cup' provide additional information, which makes the following predictions more reliable (more likely to predict `Football' and exclude `Technology, Education, Economics'). Therefore, researchers in the NLP community made great efforts in exploiting label dependency. There are two mainstream methods: the first is the explicit method, i.e., the model makes predictions based on the text and the previous predictions, and a series of studies \citep{DBLP:conf/coling/YangSLMWW18,Yang2019ADR,DBLP:journals/ipm/WangRSQD21} treated previous predictions as auxiliary information to support decision making process in different forms; the second is the implicit method, i.e., designing specific modules (e.g. attention mechanisms \citep{DBLP:conf/emnlp/XiaoHCJ19,Liu2021LabelEmbeddingBA}, graph networks \citep{DBLP:journals/ijon/LiuCLZW21,Vu2022LabelrepresentativeGC}) to capture implicit information in label dependency. Both methods demonstrated the performance gain from label dependency. However, the LD-based models suffer from two unwanted biases: 
\begin{itemize}
    \item Exposure Bias \citep{DBLP:conf/coling/YangSLMWW18}: some incorrect predictions lead to the error accumulation in the following predictions. 
    \item Stereotypes Bias \citep{DBLP:conf/acl/ZhangZYLC21}: the model tends to generate high-frequency label combinations while ignoring low-frequency label combinations. 
\end{itemize}


\begin{figure}[t]
    \centering
    \includegraphics[width=0.8\columnwidth]{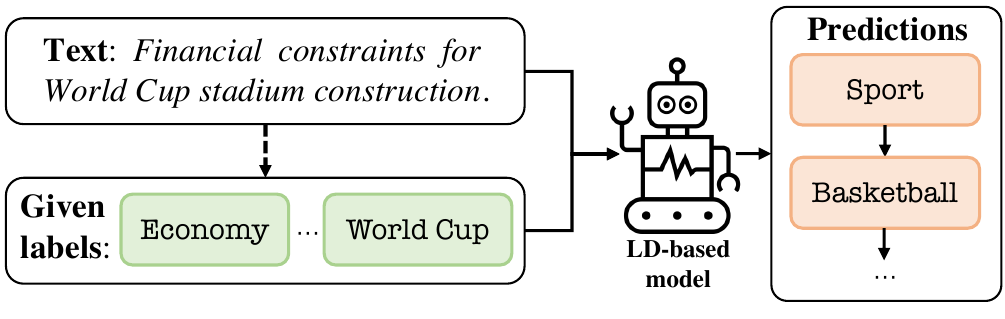}
    \caption{An example of the LD-based model dominated by label dependency. When the text and partial labels are fed to the model, the first prediction is \texttt{Sport}, as it is a high-frequency combination with \texttt{World Cup} (Stereotypes Bias), and then the incorrect prediction \texttt{Sport} leads to the following error \texttt{Basketball} (Exposure Bias). The text is ignored in this process. }
    \label{f1-1}
\end{figure}


In this study, we attribute these biases to the fact that the LD-based models may be dominated by label dependency and ignore the text, as shown in Fig. \ref{f1-1}. Through the lens of causality, MLTC's underlying mechanism is a `text $\rightarrow$ label' causality path, and label dependency should serve as auxiliary information for this path. However, label dependency, as a statistical cue \citep{Niven2019ProbingNN} between labels, has an additional `label $\rightarrow$ label' correlation shortcut. For example, if the labels `Sport, World Cup' are known in Table \ref{tab1-1}, the LD-based model can predict the label `Football' even without the text, because there is a strong correlation between them. In fact, the co-occurrence relationship between labels is sparse (in Appendix \ref{s-b}), which means that the statistical correlations between partial labels can be easily captured by the LD-based models. Since neural networks lack the ability to distinguish between correlation and causality \citep{DBLP:journals/coling/FederOSR21}, label dependency could be misused, and thus lead to unwanted correlation bias due to the correlation shortcut \citep{Shah2020ThePO}.


To avoid unwanted bias, accurate use of label dependency is crucial for LD-based models. In the human decision making process, humans usually pre-design an appropriate causal graph \citep{DBLP:journals/tkdd/YaoCLLGZ21} for a specific task, and then eliminate the correlation bias outside the causal graph to make causality-based predictions. Taking MLTC as an example, we would deliberately avoid `guessing' the predictions based only on label dependency, because our causal graph should not contain this correlation shortcut. It naturally occurs to us that if the causal graph is introduced into the LD-based model, the model could imitate the human decision making process to introspect whether the predictions are deceived by the correlation shortcut. Specifically, when the LD-based model is provided with the text and Label Information\footnote{In this paper, Label Information (LI) refers to the information extracted based on label dependency. } (LI), we employ counterfactual inference technique \citep{DBLP:conf/cvpr/NiuTZL0W21,DBLP:conf/sigir/WangF0ZC21} to simulate two states: 

\vspace{0.25cm}
\fbox{
  \parbox{0.9\textwidth}{
    \textbf{Observation}: \emph{What would the prediction be, if the model gets the text and LI?} \\
    \textbf{Intervention}: \emph{What would the prediction be, if the model could only get LI by intervening on the text?}
}}

\vspace{0.25cm}
\noindent In the observation state, the model confuses correlation and causality, while in the intervention state, the model has to rely only on the correlation shortcut to make predictions. We simulate the human de-bias process by comparing these two states. 

In this paper, we proposed a novel counterfactual framework called CounterFactual Text Classifier (CFTC) to implement the process described above. Specifically, Our CFTC was designed from two perspectives: extracting LI and using LI. We designed a novel predict-then-modify backbone in order to extract more complete and precise LI from label dependency, and employed the counterfactual intervention on the causal graph of MLTC to remove the correlation bias to enable label dependency to be used accurately. Furthermore, we extensively examined our CFTC on three datasets: AAPD \citep{DBLP:conf/coling/YangSLMWW18}, RCV1 \citep{DBLP:journals/jmlr/LewisYRL04} and Reuters-21578 \citep{Lewis1997Reuters21578TC}, and the results revealed that CFTC obtained superior performance than other baselines under the same premise of other settings. Our contributions are: 
\begin{itemize}
    \item We analyze the LD-based model's decision processes through the lens of causality and attribute the bias to the misuse of label dependency. 
    \item We employ a novel predict-then-modify backbone to extract more precise LI, and propose a counterfactual framework to block the correlation shortcut introduced by LI so that label dependency can be used accurately. 
    \item We evaluate our proposed method named CFTC on three datasets: AAPD, RCV1 and Reuters-21578, and the experimental results demonstrate the effectiveness of our CFTC. 
\end{itemize}


\section{Related Work}
\label{s2}

\subsection{Multi-Label Text Classification}
\label{s2-1}


For MLTC task, a common solution was to decompose it into multiple independent binary classification tasks, which was well known as Binary Relevance (BR) \citep{Boutell2004LearningMS}. Researchers soon realized the importance of label dependency. Label Powerset (LP) \citep{DBLP:journals/jdwm/TsoumakasK07} viewed MLTC as a multi-class classification problem by classifying data on all unique label combinations. Classifier Chains (CC) \citep{DBLP:conf/pkdd/ReadPHF09} exploited the chain rule and make predictions relying on the previous prediction. \citep{DBLP:conf/coling/YangSLMWW18,Yang2019ADR,DBLP:conf/nips/NamMKF17} viewed MLTC as a sequence generation task and utilized the Seq2Seq model as a multi-class classifier. However, both CC and Seq2Seq-based methods relied on a predefined label order. As such methods were sensitive to the label order, many studies \citep{DBLP:conf/aaai/TsaiL20} attempted to tackle the label order dependency problem. Recently, many studies proposed approaches that are not based on the Seq2Seq architecture to exploit label dependency. ML-Reasoner \citep{DBLP:journals/ipm/WangRSQD21} employed multiple rounds of predictions to obtain the final prediction. CorNet \citep{DBLP:conf/kdd/XunJSZ20} utilized BERT \citep{DBLP:conf/naacl/DevlinCLT19} and added an extra module to learn label dependency, enhance raw label predictions. LACO \citep{DBLP:conf/acl/ZhangZYLC21} and HiMatch \citep{DBLP:conf/acl/ChenMLY20} extracted label dependency using the hierarchical structure among labels and explicitly modeled the label dependency in a multi-task framework. LDGN \citep{DBLP:conf/acl/MaYZH20} learned label-specific components based on the statistical label co-occurrence in Graph Convolution Network (GCN). LELC \citep{DBLP:journals/ijon/LiuCLZW21} simplified the process of model learning by the label correlation matrix. \citep{Ozmen2022MultirelationMP} indicated that the presence or absence of each label is valuable information for MLTC. However, these methods ignore the potential bias introduced by the correlation shortcut when exploiting label dependency. 

\subsection{Counterfactual Inference}
\label{s2-2}

Counterfactual inference \citep{DBLP:journals/apin/WangLSW22} is a branch of causal inference \citep{DBLP:journals/apin/LuoZD19,DBLP:journals/tkdd/YaoCLLGZ21}, which could remove the bias in inference. Usually, counterfactual inference requires a causal graph \citep{DBLP:journals/apin/LiY20} reflecting the causal relationships between variables, and counterfactual inference means that some of the variables are fixed to values that violate the fact, thus obtaining inferences that do not fit the causal graph. A series of studies attempted to incorporate counterfactual inference into deep learning: in computer vision, \citep{DBLP:conf/cvpr/NiuTZL0W21} reduced language bias by subtracting the direct language effect from the total causal effect in Visual Question Answering, \citep{DBLP:conf/cvpr/YueWS0Z21} performed a counterfactual intervention on class attributes and obtained excellent performance in Zero-Shot Learning; in recommendation system, \citep{DBLP:conf/sigir/WangF0ZC21} used counterfactual inference to distinguish the impact of exposure features and content features, \citep{DBLP:conf/cvpr/YueWS0Z21} eliminated confounding factors in the recommendation system using counterfactual inference; in NLP, \citep{DBLP:conf/acl/0003FWMX20} reduced the document-level label bias and word-level keyword bias in text classification by counterfactual inference; \citep{DBLP:conf/aaai/WangC21} extracted causal features from the text by constructing counterfactual data, \citep{Paranjape2022RetrievalguidedCG} developed a Retrieve-Generate-Filter (RGF) technique to create counterfactual evaluation and training data with minimal human supervision. In this study, we attempt to introduce counterfactual inference into MLTC in order to eliminate the correlation bias.  


\section{Causal Analysis of MLTC}
\label{s3}

\subsection{Problem Formulation}
\label{s3-1}

MLTC studies the classification problem that each text is associated with a set of labels simultaneously. Formally, a MLTC dataset can be denoted as $D = \left\{ \mathcal{T}, \mathcal{Y} \right\}$, where $\mathcal{T}$ is the text set and $\mathcal{Y}$ is the class set. For each text $T_i \in \mathcal{T}$, it is made up of $m$ words $T_i =\{ w_{i}^1, w_{i}^2, \dots, w_{i}^m \}$, and is annotated with the corresponding label $Y_i \in \{ 0, 1 \}_{\vert L \vert }$. The target of MLTC is to learn a mapping function $F: \mathcal{T} \rightarrow \mathcal{Y}$ to minimizing the empirical risk as: 

\begin{equation}
    \min \frac{1}{N} \sum_{i=1}^N \delta (Y_i, F(T_i)), 
\end{equation}

\noindent where $\delta(\cdot)$ refers to the loss function used in the training process. 

\subsection{Causal Graph and Intervention}
\label{s3-2}



From the perspective of the LD-based model, MLTC can be formalized as a two-step process: extracting label information based on label dependency and fusing label information with text to make decisions. We notate this process as: 

\begin{equation}
    Y_{T+LI}=F(T, LI), 
    \label{e3-1}
\end{equation}

\noindent where $LI$ refers to Label Information extracted from label dependency. This format is intuitive because LI brings additional valuable information. 

\begin{figure}[t]
    \centering
    \subfigure[Human]{
    \includegraphics[height=0.209\columnwidth]{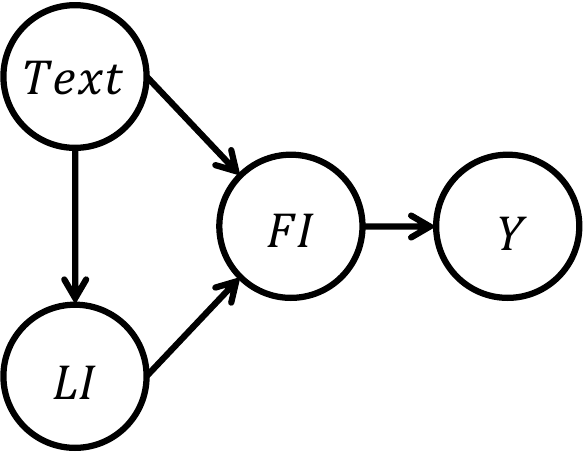} 
    \label{f3-1a}
    }
    \subfigure[LD-based Model]{
    \includegraphics[height=0.209\columnwidth]{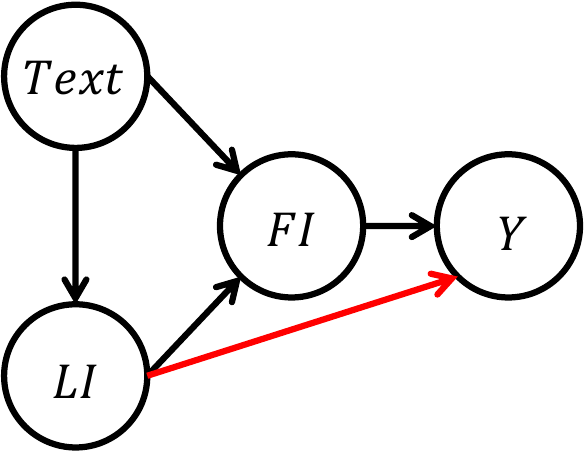} 
    \label{f3-1b}
    }
    \subfigure[Intervention]{
    \includegraphics[height=0.209\columnwidth]{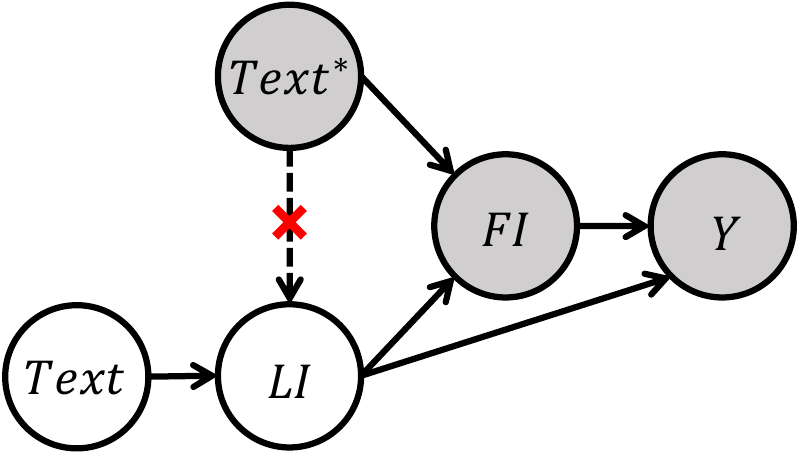}
    \label{f3-1c}
    }
    \caption{Causal graphs and Counterfactual Intervention in MLTC. $LI$ and $FI$ refer to Label Information and Fusion Information, respectively, and $Y$ is the prediction. `Grey' refers to the counterfactual part. }
    \label{f3-1}
\end{figure}

In the ideal state, the model follows the human decision making process to fuse the text and LI (Text $\rightarrow$ FI $\leftarrow$ LI) and make predictions based on the causality path (FI $\rightarrow$ Y). However, since the LD-based model cannot distinguish between causality and correlation, the neural network's preference for correlation features \citep{Du2021TowardsIA} may cause the model to be dominated by the label correlation shortcut (LI $\rightarrow$ Y), i.e., the conditional probability of the prediction collapses as: 

\begin{equation}
    p(Y \vert T, LI) \approx p(Y \vert LI), 
    \label{e3-2}
\end{equation}

\noindent which is the source of unwanted correlation bias. Following the concepts of causal inference, we construct the causal graphs of human and the LD-based model in MLTC as illustrated in  Fig. \ref{f3-1a} \& \ref{f3-1b}. The only difference between these two causal graphs is the LD-based model's predictions can be interfered with by the LI $\rightarrow$ Y shortcut (the red line in  Fig. \ref{f3-1b}). Therefore, we expect LD-based models to imitate humans by deliberately blocking this shortcut. 

However, Eq. \ref{e3-1} confuses the effect of text and LI for prediction, so we cannot distinguish whether the shortcut interferes with the model's predictions. According to the counterfactual framework \citep{DBLP:conf/cvpr/NiuTZL0W21,Wang2021DeconfoundedRF,DBLP:conf/sigir/WangF0ZC21}, we estimate the label correlation shortcut by blocking the correct text information as: 

\begin{equation}
    Y_{T^*+LI}=F(T^*, LI), 
    \label{e3-3}
\end{equation}


\noindent where $T^*$ means the counterfactual text\footnote{In this paper, we only require that $T^*$ does not contain the correct text information in $T$, rather than a semantic counterfactual text. } for the text $T$. Eq. \ref{e3-3} describes the scenario in Fig. \ref{f3-1c}: the model does not have access to the correct text information and only makes predictions based on the correct LI. This process imitates human `guessing' the predictions based only on label dependency. Thus, $F(T^*, LI)$ is a reasonable estimate of the label correlation shortcut. Then, we employ the counterfactual de-bias technique \citep{DBLP:conf/cvpr/NiuTZL0W21} to remove the label correlation bias outside the human causal graph as: 

\begin{equation}
    Y_{cd} = F(T, LI) - F(T^*, LI). 
    \label{e3-4}
\end{equation}

Intuitively, the model obtains de-biased predictions by subtracting the interference of label correlation shortcut, similar to the causal effect estimation \citep{DBLP:journals/datamine/ChengLLLLY22}. In terms of learning strategy, the counterfactual inference decouples causality and correlation in datasets based on the causal graph in Fig. \ref{f3-1a}, so the model could eliminate the label correlation bias and make causality-based predictions. 

\section{Methodology}
\label{s4}

\begin{figure*}[t]
    \centering
    \includegraphics[width=0.99\columnwidth]{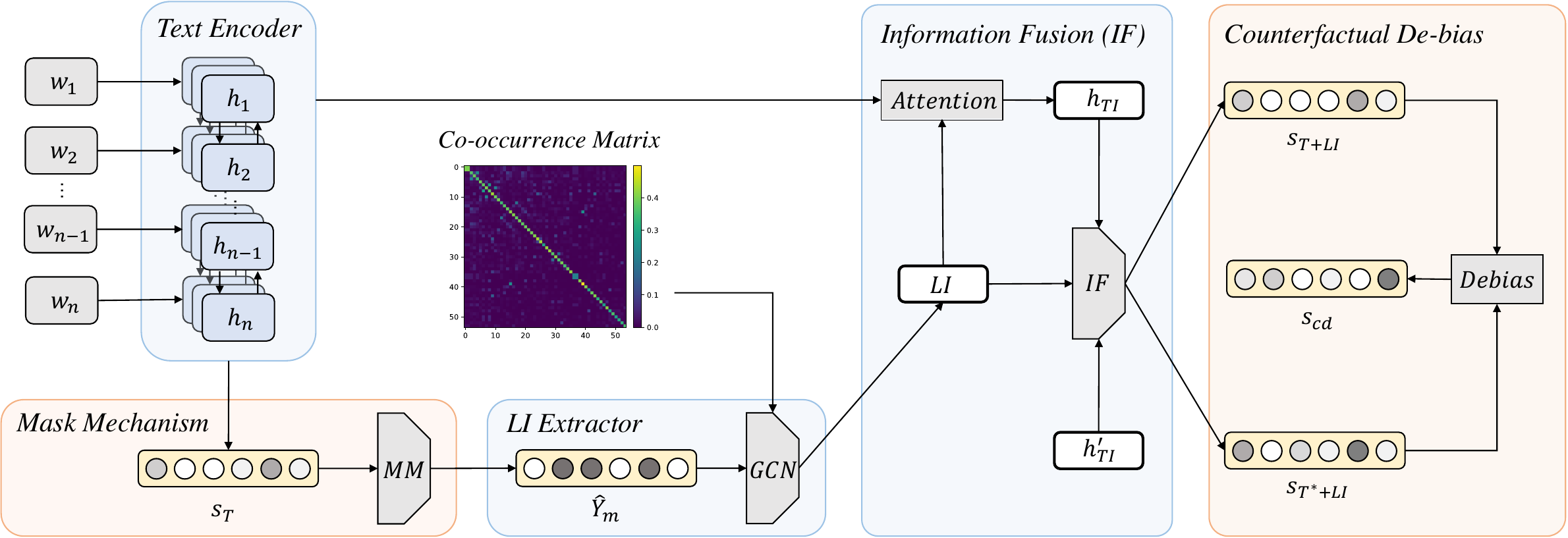}
    \caption{The architecture of the CounterFactual Text Classifier (CFTC). The predict-then-modify backbone (drawn in blue) extracts and utilizes LI based on the text and label dependency. The model eliminates the label correlation bias through Mask Mechanism (MM) and Counterfactual De-bias modules (drawn in red). }
    \label{f4-1}
\end{figure*}

To implement counterfactual de-bias process in Section \ref{s3}, we introduce our CounterFactual Text Classifier (CFTC), as shown in Fig. \ref{f4-1}. In order to make full use of label dependency and extract precise LI, we employ the predict-then-modify backbone in Section \ref{s4-1}. To remove the correlation bias from label dependency, we introduce the counterfactual de-bias in our CFTC in Section \ref{s4-2}. The training details are described in Section \ref{s4-3}. 

\subsection{Predict-Then-Modify Backbone}
\label{s4-1}

The premise of using label dependency is to extract precise LI from label dependency. However, both explicit and implicit methods of LD-based models have limitations in extracting LI: the available labels in explicit methods (e.g., Seq2Seq models) are incomplete and shallow; the label features extracted by implicit methods are not guaranteed to be pure, as text information may be mixed in. 

To overcome these limitations, we adopt the predict-then-modify backbone in our CFTC: the initial prediction $\widehat{Y}_T$ is first obtained by the text only, then LI is extracted from $\widehat{Y}_T$, the final prediction is obtained based on the text and LI. This backbone makes two complete predictions, and the latter prediction utilizes information from the previous prediction, so we name it predict-then-modify backbone. 

In this backbone, LI is extracted from the initial prediction $\widehat{Y}_T$ (Section \ref{s4-1-0}), which ensures the completeness and purity of LI; the LI extractor based on graph neural network (Section \ref{s4-1-1}) ensures that the deeper LI would not be missed; the attention mechanism (Section \ref{s4-1-2}) ensures the effective fusion of text and LI. 

\subsubsection{Text Encoder}
\label{s4-1-0}

Just like the traditional method, we employ a text encoder to transform the original text $T =\{ w^1, w^2, \dots, w^m \}$ into text information $H$. It’s worth mentioning that CFTC is encoder-agnostic and most text encoders are available for CFTC. Here, we take BiLSTM \citep{DBLP:journals/neco/HochreiterS97} as an example.

First, each word $w^i$ in the text would be converted into a word vector $e^i \in \mathbb{R}^{D_W}$, so the text would be transformed into a series of word vectors as $E = \{e^1, e^2, \dots, e^T \}$. Then, the forward hidden state $\overrightarrow{h^i} \in \mathbb{R}^{D_T}$ and the backward
hidden state $\overleftarrow{h^i} \in \mathbb{R}^{D_T}$ at step $i$ can be computed with two LSTM models from both directions as: 

\begin{equation}
    \begin{split}
        \overleftarrow{h^i} &= \overleftarrow{\text{LSTM}}(e^i, \overleftarrow{h^{i+1}}), \\
        \overrightarrow{h^i} &= \overrightarrow{\text{LSTM}}(e^i, \overrightarrow{h^{i-1}}), 
    \end{split}
\end{equation}

\noindent where $\overrightarrow{h^{i-1}}$ and $\overleftarrow{h^{i+1}}$ represent the previous hidden states from two directions, respectively. The final hidden state at step $i$ is the concatenation of two hidden states as $h^i = [\overrightarrow{h^i} \oplus \overleftarrow{h^i}] \in \mathbb{R}^{2D_T}$, where $\oplus$ denotes the concatenation operation. A series of hidden states $\{ h^1, h^2, \dots, h^m \} \in \mathbb{R}^{2D_T \times \vert m \vert }$  is considered as text information $H$ extracted from the text encoder. Finally, text information $H$ is fed into the scoring module $f_T(\cdot)$ to obtain each label's score $s_T \in \mathbb{R}^{ \vert L \vert }$ as: 

\begin{equation}
    s_T = f_T(\text{pool}(H)) = W_T \cdot \text{pool}(H) + b_T, 
    \label{e4-1}
\end{equation}

\noindent where $W_T \in \mathbb{R}^{\vert L \vert \times 2D_T}$ and $b_W \in \mathbb{R}^{\vert L \vert}$ represent the weight parameter and the bias parameter in $f_T(\cdot)$, respectively. The initial prediction $\widehat{Y}_T$ can be derived from $s_T$ by activation function $\text{Sigmoid}(\cdot)$ and the threshold $\mu$ as: 

\begin{equation}
\widehat{Y^i_T} = \begin{cases}
1, &\text{if} \ \text{Sigmoid}(s_T^i) \geq \mu; \\
0, &\text{if} \ \text{Sigmoid}(s_T^i) < \mu. 
\end{cases}
\end{equation}

\subsubsection{Graph Neural Network LI Extractor}
\label{s4-1-1}

We consider that $\widehat{Y}_T$ as a prediction is the shallowest LI, and mutual interactions between labels \citep{DBLP:conf/acl/MaYZH20} can be exploited to extract deeper LI. To capture the implied interactions of labels, we employ the label co-occurrence matrix \citep{DBLP:conf/acl/MaYZH20,DBLP:journals/ijon/LiuCLZW21} as prior knowledge and apply a graph neural network to extract deeper LI. The label co-occurrence matrix $M \in \mathbb{R}^{ \vert L \vert \times \vert L \vert }$ is the statistic of co-occurrence between labels, where $M_{ij}$ denotes the conditional probability of a text belonging to label $L_i$ when it belongs to label $L_j$. Following the normalization method in \citep{DBLP:conf/iclr/KipfW17}, the label co-occurrence matrix $M$ is normalized as: 

\begin{equation}
    \widehat{M} = D^{-\frac{1}{2}} \cdot M \cdot D^{-\frac{1}{2}}, 
    \label{e4-2}
\end{equation}

\noindent where $D$ is a diagonal degree matrix of $M$. The visualization of the label co-occurrence matrix is in Appendix \ref{s-b}. 

Specifically, we utilize GCN \citep{DBLP:conf/iclr/KipfW17} to extract deep LI from shallow $\widehat{Y}_T$. First, $\widehat{Y}_T$ needs to be embedded into LI space. \citep{Ozmen2022MultirelationMP} indicated that the presence or absence of each label is valuable information, so we design two embeddings $\{ E_i^{in},E_i^{out} \} \in \mathbb{R}^{D_L}$ for each label $L_i$ to represent whether $L_i$ appears in $\widehat{Y}_T$, and each label chooses the proper embedding based on $\widehat{Y}_T$ to compose $E_L \in \mathbb{R}^{ \vert L \vert  \times D_L}$, which is initialized to $LI_0$ as the shallowest LI. Then, $\widehat{M}$ is employed as the adjacency matrix in multi-layer GCN. Each GCN layer takes the LI of the previous GCN layer $LI_i$ as input to extract deeper $LI_{i+1}$. The layer-wise propagation rule is as follows: 

\begin{equation}
    LI_{i+1} = \sigma(\widehat{M} \cdot LI_i \cdot W_i), 
    \label{e4-3}
\end{equation}

\noindent where $\sigma(\cdot)$ denotes the ReLU activation function, and $W_i$ is the learnable parameter in the GCN layer. Suppose there are $n$ layers of GCN and $LI$ is finally represented as: 

\begin{equation}
    LI=\text{pool}(LI_n) \in \mathbb{R}^{D_L}. 
\end{equation}

The knowledge in the label co-occurrence matrix is sourced from the deterministic label distribution, so the initial LI here is embedded from the discrete prediction $\widehat{Y}_T$, rather than the continuous label-specific features \citep{DBLP:conf/emnlp/XiaoHCJ19,DBLP:conf/acl/MaYZH20}. This design prevents LI from mixing with text information, which helps our LI extractor to obtain more precise and consistent LI. 

\subsubsection{LI-Attention Information Fusion}
\label{s4-1-2}

In the predict-then-modify backbone, the final prediction is obtained based on the text and LI. Given text information $H$ and $LI$, the most common information fusion module is concatenating these two parts and feeding it to scoring function  $f_{T+LI}(\cdot)$ as: 

\begin{equation}
    s_{T+LI} = f_{T+LI}(\text{pool}(H) \oplus LI). 
    \label{e4-4}
\end{equation}

However, \citep{Chen2020HyperbolicCN} pointed out that simple feature aggregation operations (e.g. $\text{pool}(\cdot)$) would limit the model's performance because each label's related component is different in a text. Considering the semantic information of labels determine the semantic connection between labels and texts \citep{DBLP:conf/emnlp/XiaoHCJ19}, we propose the LI-attention information fusion module to capture each text’s feature. Specifically, we explicitly represent the semantic connection between $H$ and $LI$ by $LI$-$H$ attention score $A \in \mathbb{R}^{m}$ as: 

\begin{equation}
    A =\text{softmax}(H^T \cdot (W_a \cdot LI)), 
    \label{e4-5}
\end{equation}

\noindent where $W_a \in \mathbb{R}^{2D_T \times D}$ is a weight parameter. The LI-specific text information $h_{TI}$ can be obtained by a linear combination of $H$ with the help of $A$ as: 

\begin{equation}
    h_{TI} = \sum_{k=1}^m A_k h^k. 
\end{equation}

Then, the score fusing text and LI can be expressed as: 

\begin{equation}
\begin{split}
    s_{T+LI} &= f_{T+LI} \left(h_{TI} \oplus LI \right) \\
    &= W_{T+LI} \cdot \left( h_{TI} \oplus LI \right) + b_{T+LI}, 
\end{split}
\label{e4-15}
\end{equation}

\noindent where $W_{T+LI} \in \mathbb{R}^{\vert L \vert \times (2D_T + D_L)}$ and $b_{T+LI} \in \mathbb{R}^{\vert L \vert}$ represent the weight parameter and the bias parameter in $f_{T+LI}(\cdot)$, respectively. 


\subsection{Counterfactual De-bias}
\label{s4-2}


Despite the extraction and utilization of more complete and precise LI in Section \ref{s4-1}, the predict-then-modify backbone still adheres to the LD-based model's decision causal graph in Fig. \ref{f3-1b}, so the label correlation bias still exists. Therefore, we should block the label correlation shortcut to eliminate this bias, as mentioned in Section \ref{s3}. 

To measure the effect of the label correlation shortcut, we assume a counterfactual text $X^*$ for each text in Section \ref{s3}. Ideally, LI-specific counterfactual text information $h_{TI}^*$ can be obtained using a process similar to that of extracting $h_{TI}$: first, the counterfactual text information $H^*$ is extracted from $X^*$ using the text encoder in Section \ref{s4-1-0}, then, $h_{TI}^*$ is obtained by combining the existing LI through the attention module in Section \ref{s4-1-2}. However, the reality is that the counterfactual texts do not exist in most datasets. Although counterfactual data augmentation methods \citep{DBLP:conf/aaai/WangC21,Chen2021ReinforcedCD} are widely discussed, the introduction of data augmentation methods would present additional challenges, for example, additional bias may be created in the data augmentation process. 

To solve this problem, we employ a general proxy text information $h_{TI}^{'}$ as an alternative to all counterfactual text information. This method of designing proxy information for counterfactual inference is widely employed in the field of computer vision \citep{DBLP:conf/nips/TangHZ20,DBLP:conf/cvpr/NiuTZL0W21,DBLP:conf/cvpr/YangZQ021}. In this study, we extend this method to NLP problem. According to the counterfactual intervention in Fig. \ref{f3-1c}, the motivation for the intervention is to prevent the model from getting the correct text information, rather than a semantic counterfactual text. Since $h_{TI}^{'}$ is general, it would not carry any text information for a specific text, so this alternative fits the motivation. In this situation, the model can only make predictions based on the correct LI. 

Specifically, we design a trainable parameter $h_{TI}^{'} \in \mathbb{R}^{2D_T}$ to represent the proxy text information, and imitating the counterfactual intervention process in Fig. \ref{f3-1c}, $h_{TI}$ in Eq. \ref{e4-15} is replaced by $h_{TI}^{'}$, and the label correlation scores can be obtained using $f_{T+LI}(\cdot)$ based on $h_{TI}^{'}$ and $LI$ as: 

\begin{equation}
\begin{split}
    s_{T^* + LI} &= f_{T + LI} (h_{TI}^{'} \oplus LI ) \\
    &= W_{T+LI} \cdot ( h_{TI}^{'} \oplus LI ) + b_{T+LI}. 
\end{split}
\end{equation}

In the counterfactual inference, $s_{T^*+LI}$ is the estimate of the label correlation shortcut. Based on the counterfactual de-bias technique in Eq. \ref{e3-4}, we explicitly block the label correlation shortcut by subtracting $s_{T^*+LI}$ from $s_{T+LI}$, and the counterfactual de-bias score is denoted as: 

\begin{equation}
    s_{cd} = s_{T+LI} - s_{T^*+LI}. 
    \label{e4-8}
\end{equation}

Essentially, the counterfactual de-bias process provides the human decision logic to the LD-based model in the form of the causal graph, which prevents the model from making inferences based on the correlation shortcut in datasets. However, this decision logic is difficult to be learned by the model through data-driven methods. 

\subsubsection*{Mask Mechanism}


The predict-then-modify backbone, while blocking the correlation shortcut through the counterfactual de-bias technique, may cause the FI $\rightarrow$ Y path in Fig. \ref{f3-1b} to lack expressiveness. Specifically, during the training process, CFTC could extract accurate LI with the help of the ground truth, and accurate LI can then reconstruct the ground truth (Y $\rightarrow$ LI $\rightarrow$ Y). As a result, the expressiveness of the FI $\rightarrow$ Y path is likely to be weakened, because LI provides sufficient information. Although the label correlation shortcut would be blocked by counterfactual de-bias (Eq. \ref{e4-8}), weakening the expressiveness of the FI $\rightarrow$ Y path is harmful to CFTC, because it is the unique causal path in MLTC. 

To strengthen the FI $\rightarrow$ Y path, we insert the Mask Mechanism (MM) before the LI Extractor (Section \ref{s4-1-1}). Here, mask means to invert the result of whether the text matches the label or not. The purpose of the mask mechanism is to actively create uncertainties in LI so that the text information can be utilized to remove these uncertainties, which means the LD-based model has to strengthen the FI $\rightarrow$ Y path for prediction. 

We use two mask mechanisms in our CFTC: probability-based mask and random mask. In the probability-based mask, labels with low prediction confidence are more likely to be masked, and we achieve it through the Gumbel-Softmax trick \citep{DBLP:conf/iclr/JangGP17}. Specifically, we achieve this by changing the probability distribution of each label, when we get $s_T$ in Eq. \ref{e4-1}, the probability after the probability-based mask of each label is calculated as: 

\begin{equation}
\begin{split}
    \sigma_T^i &= \log (\text{Sigmoid}([s_T^i, -s_T^i])), \\
    \widehat{p_M}^i &= \text{Softmax}((\sigma_T^i + g) / \tau), 
    \label{e4-9}
\end{split}
\end{equation}

\noindent where $g \in \mathbb{R}^2$ is sampled from Gumbel$(0,1)$ distribution\footnote{The Gumbel$(0,1)$ distribution can be sampled using inverse transform sampling by drawing $u \sim \text{Uniform}(0,1)$ and computing $g = -\log (- \log (u))$. } and $\tau$ is the temperature to control the probability distribution. With the Gumbel-Softmax trick, LI has a certain probability of containing the opposite information to $s_T$, when the label's confidence is low, LI is more unreliable and the model should rely more on text information for prediction. 

After the probability-based mask, in order to increase the diversity of LI, we randomly mask a certain percentage of labels. For the selected label $L_i$, the random mask is formalized as: 

\begin{equation}
    \widehat{p_M}^i = 1 - p_T^i. 
    \label{e4-10}
\end{equation}

After the mask mechanism, we can obtain the masked initial prediction $\widehat{Y}_m$ by activation function Sigmoid($\cdot$) and the threshold $\mu$. Because of the mask mechanism, LI extracted from $\widehat{Y}_m$ carries uncertainties, and we consider that these uncertainties facilitate the text information to be fully utilized. The mask mechanism serves to weaken the correlation shortcut and strengthen the causal path, which ensures the counterfactual de-bias module in Fig. \ref{f3-1c} could work properly. 

\subsection{Training and Inference}
\label{s4-3}

In this paper, $\text{pool}(\cdot)$ is mean-pooling function, and $\text{Sigmoid}(\cdot)$ function converts the scores into probabilities of prediction $\widehat{P}$. We use Binary Cross-Entropy Loss (BCELoss) to calculate the loss of the predictions. Specifically, $Y = \{y_1,y_2,\dots,y_L \}$ are the ground-truth labels of a text, where $y_i = 0, 1$ denotes whether the text matches $L_i$ or not, and the loss can be calculated as: 

\begin{equation}
    \mathcal{L} = \frac{1}{L} \sum_{i=1}^{L} y_i \log (\widehat{p}_i) + (1-y_i) \log (1-\widehat{p}_i). 
    \label{e4-11}
\end{equation}

In addition to supervising $\{\widehat{P}_T, \widehat{P}_{T+LI}, \widehat{P}_{cd}\}$, due to the introduction of the trainable counterfactual text information $h^*_{LI}$, we also supervise $\widehat{P}_{T^*+LI}$. Therefore, the final loss is the combination of four predictions: 

\begin{equation}
    \mathcal{L} = \mathcal{L}_T + \alpha \cdot \mathcal{L}_{T+LI} + \beta \cdot \mathcal{L}_{T^*+LI} + \gamma \cdot \mathcal{L}_{cd}, 
    \label{e4-12}
\end{equation}

\noindent where $\alpha, \beta, \gamma$ are the weights of each loss, respectively. In the inference process, we use $\widehat{P}_{cd}$ as predictions because the label correlation shortcut should be blocked. 


\section{Experiments}
\label{s5}


\subsection{Datasets}
\label{s5-1}

\begin{table}[t]
    \centering
    \caption{Statistics of datasets. $N$ and $L$ denote the total number of samples and labels, respectively. $\overline{W}$ is the average number of words per sample, and $\overline{L}$ is the average number of labels per sample. }
    \fontsize{10}{10}\selectfont 
    \begin{threeparttable}
    \begin{tabular}{lccccc}
    \toprule
    \bf Dataset & $N$ & $L$ & $\overline{W}$ & $\overline{L}$ \cr
    \midrule
        \bf AAPD & 55840 & 54 & 163 & 2.41 & \cr
        \bf RCV1 & 804414 & 103 & 124 & 3.24 & \cr
        \bf Reuters-21578 & 10789 & 90 & 160 & 1.13 & \cr
    \bottomrule
    \end{tabular}
    \end{threeparttable}
    \label{tab4-1}
\end{table}

We conducted our experiments on three datasets: 

\textbf{Arxiv Academic Paper Dataset (AAPD)} was built by \citep{DBLP:conf/coling/YangSLMWW18}. It consists of abstracts and corresponding topics of papers in the field of computer science, which is organized into 54 related topics. AAPD is available in\footnote{\url{https://git.uwaterloo.ca/jimmylin/Castor-data/tree/master/datasets/AAPD}}. 

\textbf{Reuters Corpus Volume I (RCV1)} was built by \citep{DBLP:journals/jmlr/LewisYRL04}. It consists of over 80K manually categorized news made available by Reuters Ltd for research purposes, and each news is assigned multiple topics. RCV1 is available in\footnote{\url{http://www.ai.mit.edu/projects/jmlr/papers/volume5/lewis04a/lyrl2004_rcv1v2_README.htm}}. 

\textbf{Reuters-21578} was built by \citep{Lewis1997Reuters21578TC}. Its initial version contains 21578 documents and 90 categories. After eliminating the documents without categories, the final version contains 10788 documents. Reuters-21578 is available in\footnote{\url{http://www.daviddlewis.com/resources/testcollections/reuters21578/}}. 

The statistics of the datasets are listed in Table \ref{tab4-1}. 

\subsection{Evaluation Metrics}
\label{s5-2}

Following the previous works \citep{DBLP:conf/coling/YangSLMWW18,DBLP:journals/ipm/WangRSQD21}, we adopted Hamming Loss \citep{DBLP:conf/colt/SchapireS98} and Micro-$F_1$ as our main evaluation metrics. We also recorded Micro-$P$ and Micro-$R$ as secondary metrics to further analyze the experimental results. 

Hamming Loss is the fraction of labels that are incorrectly predicted, including predicted irrelevant labels and missed relevant labels. It can be computed as: 

\begin{equation}
    \text{Hamming Loss}(Y, \widehat{Y}) = \frac{1}{L} \sum_{i=1}^L  \mathbf{1}(y_i, \widehat{y}_i) ,
\end{equation}

\noindent where $\mathbf{1}(a, b)$ is the indicator function. The indicator is equal to 1 if $a=b$, otherwise it is equal to 0. 

Micro-$P$, Micro-$R$ and Micro-$F_1$ are common evaluation metrics in classification tasks. Specifically, given True Positive $TP_i$, False Positive $FP_i$, False Negative $FN_i$, and True Negative $TN_i$ for class $i$, Micro-$P$, Micro-$R$ and Micro-$F_1$ can be represented as: 

\begin{equation}
\begin{split}
    \text{Micro-}P &= \frac{\sum_{i=1}^L TP_i}{\sum_{i=1}^L TP_i + FP_i}, \\
    \text{Micro-}R &= \frac{\sum_{i=1}^L TP_i}{\sum_{i=1}^L TP_i + FN_i}, \\
    \text{Micro-}F_1 &= \frac{\sum_{i=1}^L 2TP_i}{\sum_{i=1}^L 2TP_i + FP_i + FN_i} .
\end{split}
\end{equation}

\subsection{Baselines}
\label{s5-3}


In order to verify the effectiveness of CFTC, we selected several multi-label classification algorithms as baselines. We divided the baselines into two groups based on whether label dependency is utilized. 

The first group of baselines did not use label dependency, and only focus on texts: 
\begin{itemize}
    \item \textbf{Binary Relevance (BR)} \citep{DBLP:conf/epia/GoncalvesQ03} trains a binary classifier (linear SVM) for each label, and each classifier is independent; 
    \item \textbf{Label Powerset (LP)} \citep{Boutell2004LearningMS} views MLTC as a multi-class classification problem;
    \item \textbf{CNN} \citep{DBLP:conf/emnlp/Kim14} extracts text features by multiple convolution kernels, which is a common way to extract text features; 
    \item \textbf{BiLSTM} \citep{DBLP:journals/neco/HochreiterS97} applies a Long Short-Term Memory network to extract text features, and this approach takes into account the sequential structure of the text; 
    \item \textbf{CNN-RNN} \citep{Chen2017EnsembleAO} extracts local and global text features by CNNs and RNNs; 
    \item \textbf{BERT} \citep{DBLP:conf/naacl/DevlinCLT19} applies Bidirectional Encoder Representations from Transformers to extract text features, and BERT is pre-trained on a large-scale corpus. 
\end{itemize}

The second group of baselines attempted to utilize label dependency, and except for LBA, none of these methods employed pre-trained language models: 

\begin{itemize}
    \item \textbf{Classifier Chains (CC)} \citep{DBLP:conf/pkdd/ReadPHF09} transforms the MLTC problem into a sequence of binary classification tasks; 
    \item \textbf{ML-GCN} \citep{Chen2019MultiLabelIR} captures and explores label dependency by Graph Convolutional Network and co-occurrence matrix; 
    \item \textbf{LSAN} \citep{DBLP:conf/emnlp/XiaoHCJ19} learns the label-specific text features with the help of self-attention and label-attention mechanism; 
    \item \textbf{SGM} \citep{DBLP:conf/coling/YangSLMWW18} views MLTC as a sequence generation task and utilizes seq2seq model as a multi-class classifier to use label dependency; 
    \item \textbf{ML-Reasoner (ML-R)} \citep{DBLP:journals/ipm/WangRSQD21} uses the label predictions from the previous round to utilize label dependency; 
    \item \textbf{LBA} \citep{Liu2021LabelEmbeddingBA} designs the bi-directional attentive module for the finer-grained token-level text representation and label embedding to utilize label dependency. 
\end{itemize}

\subsection{Implementation Details}
\label{s5-4}

\begin{table*}[!t]
  \centering
    \caption{Experiment results of different models on AAPD. The best performance is highlighted in  \textbf{bold}, and the best performance without the pre-trained language model is highlighted by \underline{underline}. }
  \fontsize{10}{10}\selectfont  
  \begin{threeparttable}
    \begin{tabular}{lcccc}  
    \toprule  
    \bf Models & \bf Hamming Loss $\downarrow$ & Micro-$P$ $\uparrow$ & Micro-$R$ $\uparrow$ & \bf Micro-$\bm{F_1}$ $\uparrow$ \cr
    \midrule  
    \emph{w/o LD} \cr
    BR & 0.0266 & 71.0 & 63.2 & 66.9 \cr
    LP & 0.0255 & 74.5 & 65.5 & 69.7 \cr
    CNN & 0.0259 & 72.8 & \underline{67.6} & 70.0 \cr 
    BiLSTM & 0.0254 & 74.3 & 67.2 & 70.3 \cr
    CNN-RNN & 0.0261 & 72.6 &66.9 & 69.7 \cr
    BERT & 0.0230 & 79.1 & 66.3 & 72.2 \cr
    \midrule
    \emph{LD-based} \cr

    CC & 0.0256 & 75.8 & 62.9 & 66.8 \cr
    ML-GCN & 0.0247 & \underline{78.9} & 61.3 & 69.0 \cr
    LSAN & 0.0246 & 75.2 & 67.5 & 70.9 \cr
    SGM & 0.0251 & 74.8 & 67.5 & 71.0 \cr 
    ML-R & 0.0255 & 74.6 & 65.5 & 69.8 \cr
    LBA & 0.0228 & 78.8 & 67.0 & 72.1 \cr
    \midrule
    $\text{CFTC}_{BiLSTM}$ & \underline{0.0237} & 77.0 & 66.6 & \underline{71.4} \cr
    $\text{CFTC}_{BERT}$ & \bf0.0222 & \bf79.3 & \bf68.4 & \bf73.4 \cr
    \bottomrule  
    \end{tabular}
    \end{threeparttable}
    \label{tab6-1}
\end{table*}

BiLSTM and BERT were employed as the encoders in our CFTC to extract text features, respectively. We chose a 3-layer GCN to transform the initial prediction into LI, and the hidden size of GCNs was set to 300. The word embedding dimension was 300 for BiLSTM and 768 for BERT. We set the batch size to 64 (16 for BERT), and the learning rate of the Adam optimizer to 1e-4 (5e-5 for BERT). After training models for 50 epochs (10 epochs for BERT), we selected the best model on the training set for testing. Since the text encoder part served to extract text features and was prone to overfitting, we optimized the encoder by $\mathcal{L}_T$ and optimized the decoder by $\mathcal{L}_{T+LI}$, $\mathcal{L}_{T^*+LI}$ and $\mathcal{L}_{cd}$, or pre-trained the encoder and froze the parameters on training. In this paper, we masked 5\% of labels in the mask mechanism, the threshold of probability was set to 0.5, and we set the weight $\alpha=\beta=0.1$ and $\gamma=1.0$ in the training process. 

\section{Results and Analysis}
\label{s6}

This section is mainly about the performance of CFTC. We reported the main experimental results of CFTC and other baselines on three datasets (Section \ref{s6-1}). Besides, we validated the contribution of each module in CFTC through ablation experiments (Section \ref{s6-2}). To demonstrate the effectiveness of the counterfactual de-bias technique, we compared the difference in label co-occurrence frequencies before and after counterfactual de-bias in CFTC (Section \ref{s6-3}). We showed the advantages of CFTC in eliminating the label correlation bias through case study (Section \ref{s6-4}). 

\begin{table*}[t]
  \centering
    \caption{Experiment results of different models on RCV1. The best performance is highlighted in \textbf{bold}, and the best performance without the pre-trained language model is highlighted by \underline{underline}. }
  \fontsize{10}{10}\selectfont  
  \begin{threeparttable}
    \begin{tabular}{lcccc}  
    \toprule
    \bf Models & \bf Hamming Loss $\downarrow$ & Micro-$P$ $\uparrow$ & Micro-$R$ $\uparrow$ & \bf Micro-$\bm{F_1}$ $\uparrow$ \cr
    \midrule  
    \emph{w/o LD} \cr
    BR & 0.0086 & 90.4 & 81.6 & 85.8 \cr
    LP & 0.0087 & 89.6 & 82.4 & 85.8 \cr
    CNN & 0.0083 & 88.1 & 85.1 & 86.6 \cr 
    BiLSTM & 0.0079 & 89.0 & 85.1 & 87.0 \cr
    CNN-RNN & 0.0087 & 87.8 & 83.8 & 85.8 \cr
    BERT & 0.0071 & 90.5 & 86.5 & 88.5  \cr
    \midrule
    \emph{LD-based} \cr

    CC & 0.0087 & 88.7 & 82.8 & 85.7 \cr
    ML-GCN & 0.0080 & 88.0 & 86.1 & 87.0 \cr
    LSAN & 0.0079 & \bf \underline{91.3} & 82.5 & 86.7 \cr
    SGM & 0.0079 & 88.5 & 86.0 & 87.2 \cr 
    ML-R & 0.0079 & 89.7 & 84.5 & 87.0 \cr
    LBA & 0.0073 & 90.0 & 85.9 & 88.0 \cr
    \midrule
    $\text{CFTC}_{BiLSTM}$ & \underline{0.0074} & 89.3 & \underline{86.1} & \underline{88.0} \cr
    $\text{CFTC}_{BERT}$ & \bf 0.0068 & 90.5 & \bf 87.4 & \bf 88.9 \cr
    \bottomrule  
    \end{tabular}
    \end{threeparttable}
    \label{tab6-2}
\end{table*}

\begin{table*}[t]
  \centering
    \caption{Experiment results of different models on Reuters-21578. The best performance is highlighted in \textbf{bold}, and the best performance without the pre-trained language model is highlighted by \underline{underline}. }
  \fontsize{10}{10}\selectfont  
  \begin{threeparttable}
    \begin{tabular}{lcccc}  
    \toprule  
    \bf Models & \bf Hamming Loss $\downarrow$ & Micro-$P$ $\uparrow$ & Micro-$R$ $\uparrow$ & \bf Micro-$\bm{F_1}$ $\uparrow$ \cr
    \midrule  
    \emph{w/o LD} \cr
    BR & 0.0049 & 86.2 & 78.8 & 82.3 \cr
    LP & 0.0054 & 78.7 & 81.0 & 79.8 \cr
    CNN & 0.0038 & 89.0 & \underline{82.3} & 85.5 \cr
    BiLSTM & 0.0040 & 90.4 & 78.4 & 84.0 \cr
    CNN-RNN & 0.0038 & 90.3 & 81.3 & 85.5 \cr
    BERT & 0.0031 & \bf 93.4 & 83.3 & 88.0 \cr
    \midrule
    \emph{LD-based} \cr
    CC & 0.0045 & 85.2 & 81.7 & 83.4 \cr
    ML-GCN & 0.0043 & 86.0 & 81.8 & 83.5 \cr
    LSAN & 0.0041 & 87.4 & 82.1 & 84.7 \cr
    SGM & 0.0052 & 80.7 & 75.9 & 78.8 \cr
    ML-R & 0.0041 & 91.3 & 77.5 & 83.8 \cr
    LBA & 0.0030 & 92.7 & 86.5 & 88.5 \cr
    \midrule
    $\text{CFTC}_{BiLSTM}$ & \underline{0.0037} & \underline{91.4} & 81.5 & \underline{86.2} \cr
    $\text{CFTC}_{BERT}$ & \bf 0.0029 & 91.7 & \bf 87.8 & \bf 88.8 \cr
    \bottomrule  
    \end{tabular}
    \end{threeparttable}
    \label{tab6-a}
\end{table*}

\subsection{Main Results}
\label{s6-1}

We compared the performance of CFTC as well as other compared baselines on AAPD, RCV1 and Reuters-21578. It can be observed that CFTC outperformed all other baselines in most metrics, and the results confirmed the effectiveness of CFTC. 

\subsubsection*{Results on AAPD}
As shown in Table \ref{tab6-1}, our CFTC outperformed all other baselines by a significant margin on AAPD dataset. In particular, CFTC with BERT performed best: Hamming Loss is 0.0222, which is a 3.5\% improvement over the best result in baselines, and Micro-$F_1$ also received a 1.2\% (absolute) improvement. In addition, our CFTC with two text encoders (BiLSTM, BERT) outperformed the corresponding encoder baselines, the results showed that our CFTC is effective for different text encoders. 

\subsubsection*{Results on RCV1}

Compared to AAPD dataset, the performance difference between models on RCV1 was reduced, but our CFTC continued to perform better than other baselines on main metrics. As shown in Table \ref{tab6-2}, CFTC with BiLSTM exceeded the baseline BiLSTM by 6.3\% on Hamming Loss and by 1.0\% (absolute) on Micro-$F_1$, and CFTC with BERT beat the baseline BERT by 4.2\% on Hamming Loss and by 0.4\% (absolute) on Micro-$F_1$. 

\subsubsection*{Results on Reuters-21578}

As shown in Table \ref{tab6-a}, our CFTC achieved better performance on Reuters-21578 dataset. Compared with the baseline BiLSTM, CFTC with BiLSTM achieved a 7.5\% improvement on Hamming Loss and a 2.2\% (absolute) improvement on Micro-$F_1$. Also, CFTC with BERT also defeated the baseline BERT, leading by 6.5\% on Hamming Loss and by 0.8\% (absolute) on Micro-$F_1$. 


\subsection{Ablation Study}
\label{s6-2}

\begin{table}[t]
  \centering
    \caption{Ablation analysis on main mechanisms of our framework on AAPD and RCV1. $\backslash$ denotes the removing operation. MM and CD denote the Mask Mechanism and Counterfactual De-bias, respectively}. 
  \begin{threeparttable}
  \resizebox{11.9cm}{!}{
    \begin{tabular}{lcccc}
    \toprule
    \multirow{2}{*}{\bf Models} & 
    \multicolumn{2}{c}{\bf AAPD}& \multicolumn{2}{c}{\bf RCV1} \cr
    \cmidrule(lr){2-3} \cmidrule(lr){4-5}
    & \bf Hamming Loss $\downarrow$ & \bf Micro-$\bm{F_1}$ $\uparrow$ & \bf Hamming Loss $\downarrow$ & \bf Micro-$\bm{F_1}$ $\uparrow$ \cr
    \midrule
    $\text{CFTC}_{BiLSTM}$ & \bf0.0237 & \bf71.4 & \bf0.0074 & \bf88.0 \cr
    BiLSTM & 0.0254 & 70.3 & 0.0079 & 87.0 \cr
    \midrule
    $\backslash$ MM & 0.0241 & 71.1 & 0.0076 & 87.6 \cr
    $\backslash$ CD & 0.0242 & 70.9 & 0.0077 & 87.4 \cr
    $\backslash$ MM\&CD & 0.0249 & 70.8 & 0.0079 & 87.1 \cr
    \bottomrule
    \end{tabular}
    }
    \end{threeparttable}
    \label{tab6-3}
\end{table}

We investigated the independent impact of each module in our proposed CFTC. The results are reported in Table \ref{tab6-3}. When we only employed the predict-then-modify backbone (\verb|\|MM\&CD), the model performed slightly better than BiLSTM, which illustrated the role of the precise LI. When we removed the Mask Mechanism module (\verb|\|MM) and Counterfactual De-bias module (\verb|\|CD), the model performance decreased significantly compared to $\text{CFTC}_{BiLSTM}$. This showed the effectiveness of both modules in the counterfactual de-bias part. 

Further, CFTC was more effective on AAPD dataset compared to RCV1 dataset. This result implied that the LD-based model was more susceptible to the correlation shortcut on smaller datasets, and as the size of the datasets increased, the model became more precise in mining the underlying mechanisms of the task, thus reducing the reliance on the correlation shortcut. 

\subsection{Label Co-occurrence Frequency}
\label{s6-3}

\begin{figure}[t]
    \centering
    \subfigure{
    \includegraphics[height=0.28\columnwidth]{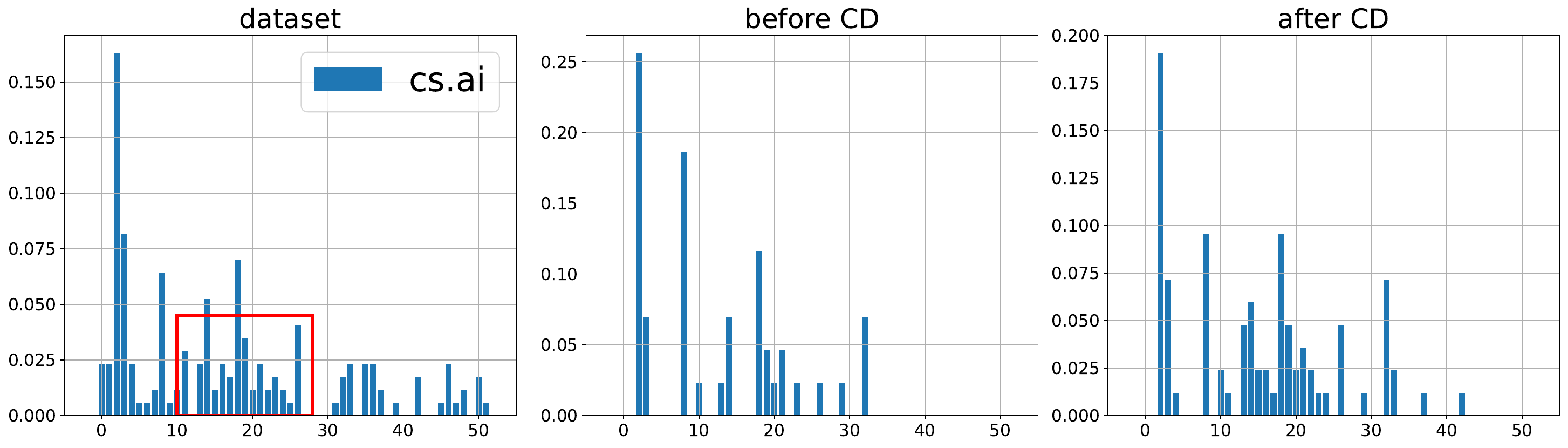}
    \label{f6-1a}
    }
    \subfigure{
    \includegraphics[height=0.28\columnwidth]{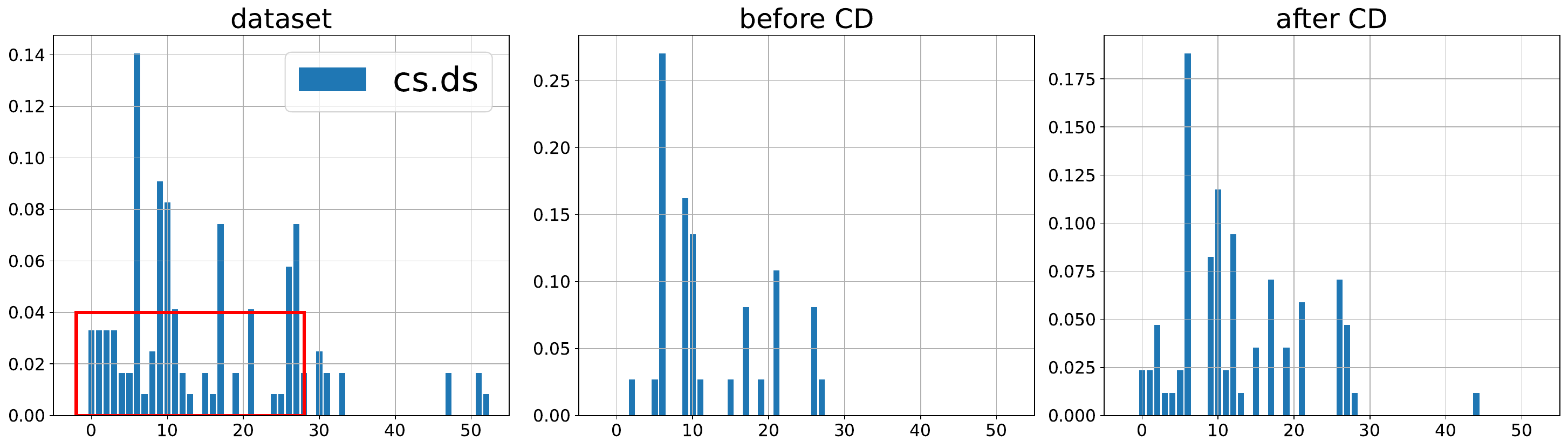} 
    \label{f6-1b}
    }
    \subfigure{
    \includegraphics[height=0.28\columnwidth]{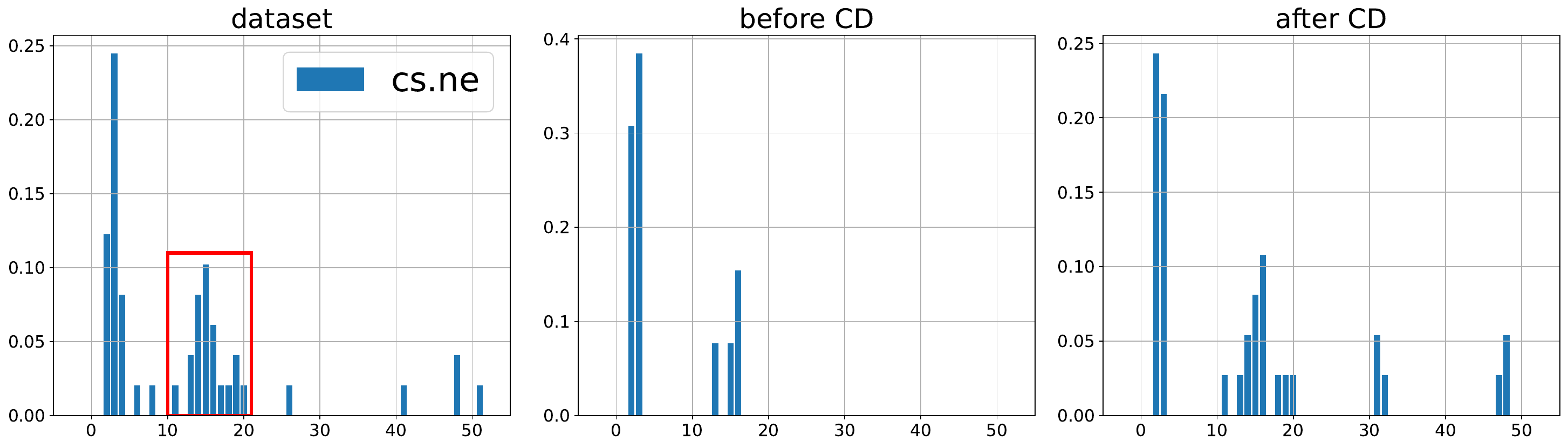}
    \label{f6-1c}
    }
    \caption{Co-occurrence frequency of some labels on AAPD dataset. The first column is the true co-occurrence frequency in the dataset, and the second and third columns are the co-occurrence frequencies of the predictions before and after Counterfactual De-bias (CD). Significant differences are boxed in \textcolor{red}{red}. }
    \label{f6-1}
\end{figure}

A notable manifestation of the label correlation bias is the LD-based model's tendency to generate high-frequency label combinations and ignore low-frequency label combinations. Therefore, the label co-occurrence frequency could, to some extent, reflect the effect of the label correlation bias. 

To demonstrate the effectiveness of the counterfactual de-bias technique in CFTC, we compared the co-occurrence frequencies of some labels (`cs.ai', `cs.ds', `cs.ne') before and after counterfactual de-bias in Fig. \ref{f6-1}. The first column showed the true co-occurrence frequency in AAPD. The second and third columns showed the co-occurrence frequencies of the predictions before and after counterfactual de-bias. It can be seen that the predictions before de-bias tended to generate high-frequency label combinations, and a large number of low-frequency label combinations were not captured. This reflected that the LD-based model was heavily influenced by the label correlation bias. In contrast, the predictions after counterfactual de-bias showed higher similarity to the true co-occurrence frequency, which reflected the effect of the counterfactual de-bias technique in eliminating the label correlation bias. 

\begin{table}[t]
  \centering
  \caption{The predictions of CFTC with different LI. We obtain different LI by changing `\textbf{Given Labels}'. \textcolor{red}{\ding{172}}, \textcolor{red}{\ding{173}}, \textcolor{blue}{\ding{174}}, \textcolor{blue}{\ding{175}} refer to \textcolor{red}{cs.lg}, \textcolor{red}{st.ml}, \textcolor{blue}{cs.it}, \textcolor{blue}{ma.it}, respectively. $\varnothing$ means empty set. }
  \begin{tabular}{ cccc }
    \toprule
    \multicolumn{4}{l}{\textbf{Ground Truth}: \textcolor{red}{\ding{172}}, \textcolor{red}{\ding{173}} \quad $Y_{T}$: \textcolor{red}{\ding{172}}, \textcolor{red}{\ding{173}}, \textcolor{blue}{\ding{174}}, \textcolor{blue}{\ding{175}} }\\
    \midrule
    \bf Given Labels & $\bm{Y}_{T^*+LI}$ & $\bm{Y}_{T+LI}$ & $\bm{Y}_{cd}$ \\
    \midrule
    \textcolor{red}{\ding{172}}, \textcolor{red}{\ding{173}}, \textcolor{blue}{\ding{174}}, \textcolor{blue}{\ding{175}}  & \textcolor{red}{\ding{172}}, \textcolor{blue}{\ding{174}}, \textcolor{blue}{\ding{175}} & \textcolor{red}{\ding{172}}, \textcolor{red}{\ding{173}}, \textcolor{blue}{\ding{174}}, \textcolor{blue}{\ding{175}} & \textcolor{red}{\ding{172}}, \textcolor{red}{\ding{173}} \\
    \textcolor{red}{\ding{172}}, \textcolor{red}{\ding{173}} & \textcolor{red}{\ding{172}}, \textcolor{red}{\ding{173}} & \textcolor{red}{\ding{172}}, \textcolor{red}{\ding{173}} & \textcolor{red}{\ding{172}}, \textcolor{red}{\ding{173}} \\
    \textcolor{red}{\ding{172}}, \textcolor{blue}{\ding{175}} & \textcolor{red}{\ding{172}} & \textcolor{red}{\ding{172}}, \textcolor{red}{\ding{173}} & \textcolor{red}{\ding{172}}, \textcolor{red}{\ding{173}} \\
    $\varnothing$ & $\varnothing$ & \textcolor{red}{\ding{173}} & \textcolor{red}{\ding{172}}, \textcolor{red}{\ding{173}} \\
    \textcolor{blue}{\ding{174}}, \textcolor{blue}{\ding{175}} & \textcolor{blue}{\ding{174}}, \textcolor{blue}{\ding{175}} & \textcolor{blue}{\ding{174}}, \textcolor{blue}{\ding{175}} & \textcolor{red}{\ding{172}}, \textcolor{red}{\ding{173}}, \textcolor{blue}{\ding{174}}, \textcolor{blue}{\ding{175}} \\
    \bottomrule
  \end{tabular}
  \label{tab6-5}
\end{table}

\begin{figure}
    \centering
    \includegraphics[width=0.4\columnwidth]{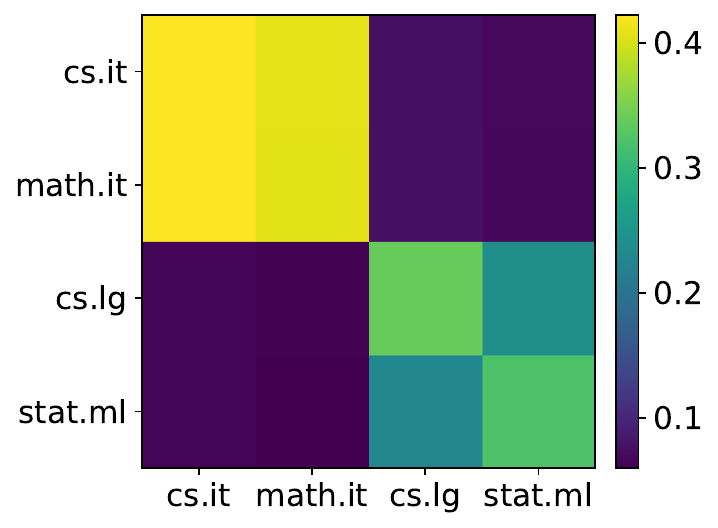}
    \caption{The visualization of the normalized label co-occurrence matrix of the four mentioned labels. The lighter the color, the higher the frequency of co-occurrence between labels}
    \label{f6-2}
\end{figure}

\subsection{Case Study}
\label{s6-4}

\subsubsection{Different Label Information}
\label{s6-4-1}

To demonstrate that CFTC is able to use LI accurately, we selected a specific text and compared predictions in the case of different LI in Table \ref{tab6-5}. The text can be found in Appendix \ref{s-a}. We obtained different LI by changing the given labels fed to the LI extractor (Although the given labels in training process are $Y_{T}$, CFTC supports arbitrarily changing the given labels in testing process). 

Row 1 was the prediction without changing LI (Given Labels was $Y_{T}$), compared to ground truth (\textcolor{red}{\ding{172}}, \textcolor{red}{\ding{173}}), $Y_{T}$ had two more error labels (\textcolor{blue}{\ding{174}}, \textcolor{blue}{\ding{175}}), and the error labels were retained in $Y_{T+LI}$, which reflected the effect of LI on $Y_{T+LI}$. However, after counterfactual de-bias, the error labels were corrected in $Y_{cd}$. To compare the effect of LI on predictions, we changed LI by selecting a portion of the labels in $Y_{T}$ as Given Labels and compared the three predictions $Y_{T^*+LI}$, $Y_{T+LI}$, $Y_{cd}$. To help readers understand the correlation between these labels, we visualized the normalized label co-occurrence matrix of these four labels in Fig. \ref{f6-2}. When feeding the correct LI to CFTC (row 2), all predictions were correct, which was in line with our expectations. Further, when LI was partially correct (row 3), $Y_{T+LI}$ and $Y_{cd}$ were both correct, while $Y_{T^*+LI}$ was incorrect, which indicated that the text was useful for predicting $Y_{T+LI}$ and $Y_{cd}$. However, when LI was incorrect, $Y_{cd}$ would be much more accurate than $Y_{T+LI}$. More specifically, missing LI (row 4) caused $Y_{T+LI}$ to be incomplete, while $Y_{cd}$ was correct, and when LI was completely incorrect (row 5), $Y_{T+LI}$ got the completely incorrect prediction consistent with Given Labels, while $Y_{cd}$ contained both the correct labels and Given Labels. This result illustrated that $Y_{T+LI}$ tended to make predictions dominated by LI, while $Y_{cd}$ would fuse text and LI to make causality-based predictions. In fact, this is evidence that CFTC has succeeded in blocking the label correlation shortcut.  

\subsubsection{Different Models}
\label{s6-4-2}

\begin{table*}[t]
  \centering  
  \caption{Four examples of labels generated by different models on AAPD dataset. The same/different colored labels indicate that they are highly correlated/uncorrelated. }
  \begin{threeparttable}
    \resizebox{11.7cm}{!}{
    \begin{tabular}{lcccc}  
    \toprule
    {\bf \#} & {\bf Ground Truth} & {$\bm{Y}_{BiLSTM}$} & {$\bm{Y}_{SGM}$} & {$\bm{Y}_{CFTC}$} \cr
    \midrule
    1 & \textcolor{red}{cs.it}, \textcolor{red}{ma.it} & \textcolor{red}{cs.it}, \textcolor{red}{ma.it}, \textcolor{blue}{cs.ni} & \textcolor{red}{cs.it}, \textcolor{red}{ma.it} & \textcolor{red}{cs.it}, \textcolor{red}{ma.it} \cr 
    2 & \textcolor{red}{cs.ai}, \textcolor{blue}{cs.pl} & \textcolor{red}{cs.ai}, \textcolor{blue}{cs.pl} & \textcolor{red}{cs.ai}, \textcolor{blue}{cs.lo}, \textcolor{blue}{cs.pl} & \textcolor{red}{cs.ai}, \textcolor{blue}{cs.pl}  \cr 
    3 & \textcolor{red}{cs.it}, \textcolor{red}{ma.it}, \textcolor{blue}{ma.oc}, ma.pr & \textcolor{red}{cs.it}, \textcolor{red}{ma.it}, \textcolor{blue}{ma.oc}, ma.pr & \textcolor{red}{cs.it}, \textcolor{red}{ma.it} & \textcolor{red}{cs.it}, \textcolor{red}{ma.it}, \textcolor{blue}{ma.oc}, ma.pr \cr 
    4 & \textcolor{red}{cs.cl}, \textcolor{blue}{cs.ir} & \textcolor{red}{cs.cl}, \textcolor{blue}{cs.ir}, \textcolor{red}{cmp-lg} & \textcolor{red}{cs.cl}, \textcolor{red}{cmp-lg} & \textcolor{red}{cs.cl}, \textcolor{blue}{cs.ir} \cr 
    \bottomrule  
    \end{tabular}
    }
    \end{threeparttable}
    \label{tab6-4}
\end{table*}

We selected several examples and showed the prediction results of different models for comparison in Table \ref{tab6-4}, where BiLSTM did not make use of label information and SGM did. All texts can be found in Appendix \ref{s-a}. The experimental results illustrated that CFTC outperformed other two models in these cases. Meanwhile, we could find the effect of the label correlation bias on the LD-based models. As shown in Table \ref{tab6-4}, in most cases, ground truths were correlated with each other, and more accurate predictions can often be obtained by the LD-based model (row 1). However, when ground truths were uncorrelated (rows 2, 3, 4), the LD-based model tended to predict the incorrect labels with high correlation ($Y_{SGM}$ in rows 2, 4) or miss correct labels with low correlation ($Y_{SGM}$ in rows 3, 4) due to the presence of the label correlation shortcut, while the model without the label dependency performed relatively better ($Y_{BiLSTM}$ in rows 2, 3, 4). Our CFTC could use the label dependency accurately, so the performance of CFTC outperformed the other two baselines in all cases. 

\section{Conclusion}

In this paper, we attributed the bias of LD-based models to the label correlation shortcut. To avoid this unwanted bias, we designed a counterfactual framework with the predict-then-modify backbone named CFTC to obtain causality-based predictions. The experimental results showed that CFTC effectively blocked the label correlation shortcut and achieved competitive performance. In the future, we hope that our proposed CFTC can be applied in more NLP scenarios and help deep learning models to better grasp the underlying mechanisms of NLP tasks. 

\subsubsection*{Competing interests}

This work was supported by the Shanghai Municipal Science and Technology Major Project (2021SHZDZX0102), and the Shanghai Science and Technology Innovation Action Plan (20511102600).

\begin{appendices}

\section{Text Contents in Analysis}
\label{s-a}

In this paper, we selected several representative samples from AAPD \citep{DBLP:conf/coling/YangSLMWW18} to analyze the model performance. We showed these contents in Table \ref{tab1} \& \ref{tab2}. AAPD contained the abstract and the corresponding subjects of 55840 papers in the computer science field from the website. 

By comparing the predictions obtained from these samples, we verified that there was unwanted bias in the LD-based model due to the label correlation shortcut, and our CFTC could alleviate this bias and obtain causality-based predictions. 

\begin{table*}[ht]
    \centering
    \caption{The text content of the samples selected in Section \ref{s6-4-1}. }
    \fontsize{5}{5}\selectfont 
    \begin{threeparttable}
    \begin{tabularx}{\textwidth}{lX}
    \toprule
    \bf \# & \bf Text \cr
    \midrule
    1 & This paper describes a simple framework for structured sparse recovery based on convex optimization. We show that many structured sparsity models can be naturally represented by linear matrix inequalities on the support of the unknown parameters, where the constraint matrix has a totally unimodular (TU) structure. For such structured models, tight convex relaxations can be obtained in polynomial time via linear programming. Our modeling framework unifies the prevalent structured sparsity norms in the literature, introduces new interesting ones, and renders their tightness and tractability arguments transparent.  \cr
    \bottomrule
    \end{tabularx}
    \end{threeparttable}
    \label{tab1}
\end{table*}

\begin{table*}[ht]
    \centering
    \caption{The text content of the samples selected in Section \ref{s6-4-2}. }  
    \fontsize{5}{5}\selectfont 
    \begin{threeparttable}
    \begin{tabularx}{\textwidth}{lX}
    \toprule
    \bf \# & \bf Text \cr
    \midrule
    1 & We consider a scenario where a monitor is interested in being up to date with respect to the status of some system which is not directly accessible to this monitor. However, we assume a source node has access to the status and can send status updates as packets to the monitor through a communication system. We also assume that the status updates are generated randomly as a Poisson process. The source node can manage the packet transmission to minimize the age of information at the destination node, which is defined as the time elapsed since the last successfully transmitted update was generated at the source. We use queuing theory to model the source-destination link and we assume that the time to successfully transmit a packet is a gamma distributed service time. We consider two packet management schemes: LCFS (Last Come First Served) with preemption and LCFS without preemption. We compute and analyze the average age and the average peak age of information under these assumptions. Moreover, we extend these results to the case where the service time is deterministic. \cr
    \midrule
    2 & The most advanced implementation of adaptive constraint processing with Constraint Handling Rules (CHR) allows the application of intelligent search strategies to solve Constraint Satisfaction Problems (CSP). This presentation compares an improved version of conflict-directed backjumping and two variants of dynamic backtracking with respect to chronological backtracking on some of the AIM instances which are a benchmark set of random 3-SAT problems. A CHR implementation of a Boolean constraint solver combined with these different search strategies in Java is thus being compared with a CHR implementation of the same Boolean constraint solver combined with chronological backtracking in SICStus Prolog. This comparison shows that the addition of ``intelligence'' to the search process may reduce the number of search steps dramatically. Furthermore, the runtime of their Java implementations is in most cases faster than the implementations of chronological backtracking. More specifically, conflict-directed backjumping is even faster than the SICStus Prolog implementation of chronological backtracking, although our Java implementation of CHR lacks the optimisations made in the SICStus Prolog system. To appear in Theory and Practice of Logic Programming (TPLP). \cr
    \midrule
    3 & In this paper we look at isometry properties of random matrices. During the last decade these properties gained a lot attention in a field called compressed sensing in first place due to their initial use in [7, 8]. Namely, in [7, 8] these quantities were used as a critical tool in providing a rigorous analysis of $l_1$ optimization’s ability to solve an under-determined system of linear equations with sparse solutions. In such a framework a particular type of isometry, called restricted isometry, plays a key role. One then typically introduces a couple of quantities, called upper and lower restricted isometry constants to characterize the isometry properties of random matrices. Those constants are then usually viewed as mathematical objects of interest and their a precise characterization is desirable. The first estimates of these quantities within compressed sensing were given in [7, 8]. As the need for precisely estimating them grew further a finer improvements of these initial estimates were obtained in e.g. [2, 4]. These are typically obtained through a combination of union-bounding strategy and powerful tail estimates of extreme eigenvalues of Wishart (Gaussian) matrices (see, e.g. [19]). In this paper we attempt to circumvent such an approach and provide an alternative way to obtain similar estimates. \cr
    \midrule
    4 & This article evaluates the performance of two techniques for query reformulation in a system for information retrieval, namely, the concept based and the pseudo relevance feedback reformulation. The experiments performed on a corpus of Arabic text have allowed us to compare the contribution of these two reformulation techniques in improving the performance of an information retrieval system for Arabic texts.  \cr
    \bottomrule
    \end{tabularx}
    \end{threeparttable}
    \label{tab2}
\end{table*}

\section{Label Co-occurrence Matrix}
\label{s-b}

\begin{figure}[t]
    \centering
    \subfigure[AAPD]{
    \includegraphics[height=0.25\columnwidth]{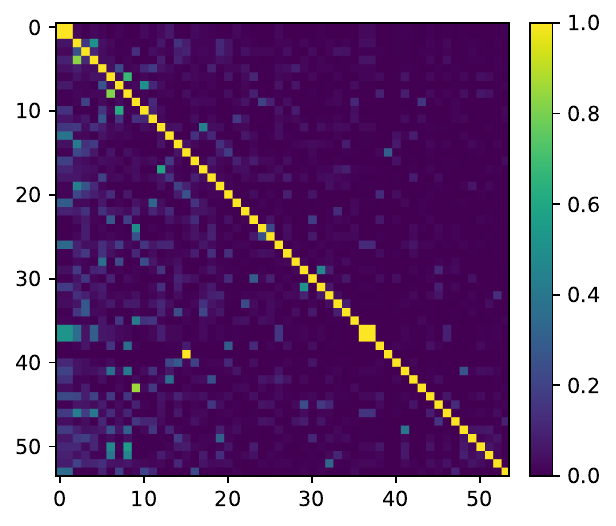} 
    \label{f1a}
    }
    \subfigure[RCV1]{
    \includegraphics[height=0.25\columnwidth]{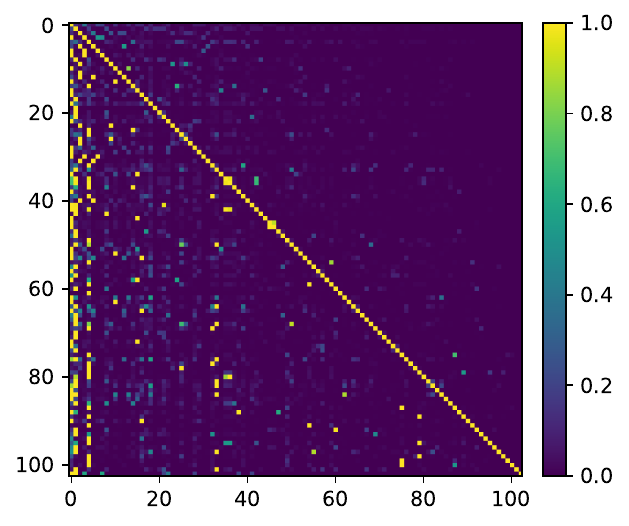}
    \label{f1b}
    }
    \subfigure[Reuters-21578]{
    \includegraphics[height=0.25\columnwidth]{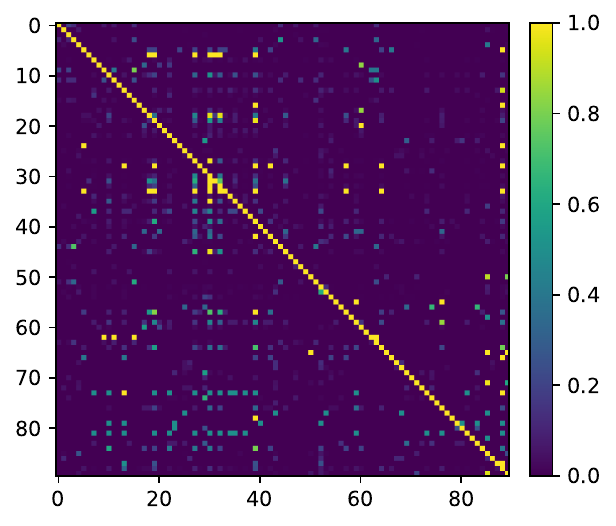}
    \label{f1c}
    }
    \caption{The label co-occurrence matrix of AAPD, RCV1 and Reuters-21578. }
    \label{f1}
\end{figure}

To capture the implied interactions of labels, we employed the label co-occurrence matrix \citep{DBLP:conf/acl/MaYZH20,DBLP:journals/ijon/LiuCLZW21} as prior knowledge and applied a graph neural network to extract deeper label information. The label co-occurrence matrix $A \in \mathbb{R}^{\vert L\vert \times\vert L\vert }$ is the statistic of co-occurrence between labels, where $A_{ij}$ denotes the conditional probability of a text belonging to label $L_i$ when it belongs to label $L_j$. We counted the label co-occurrence matrices of AAPD, RCV1 and Reuters-21578, and visualized them as in Fig. \ref{f1}. 

The results showed the existence of sparse co-occurrence relationships between the labels, and this particular relationship can provide additional information to the model in Multi-Label Text Classification tasks. 





\end{appendices}


\clearpage

\bibliography{reference}


\end{document}